%% file: main.tex
\documentclass[runningheads]{llncs}
\usepackage{graphicx}
\newif\ifanonymous
\anonymousfalse
\newcommand{\anonymize}[2]{%
  \ifanonymous
    #2 % Display the second input when the flag is set
  \else
    #1 % Display the first input otherwise
  \fi
}

\usepackage{array,booktabs}
\newcommand {\otoprule}{\midrule [\heavyrulewidth]}
\newcolumntype {+}{ >{\global\let\currentrowstyle\relax}}
\newcolumntype {^}{ >{\currentrowstyle }}
 \newcommand {\rowstyle}[1]{\gdef\currentrowstyle{#1} %
 #1\ignorespaces
 }
\newcommand{\tabhead}{\rowstyle{\bfseries}}
\usepackage{tablefootnote}
\newcommand{\tablefootnotemark}[1]{\textsuperscript{\ref{#1}}}
\usepackage{fancyhdr}

\usepackage{amssymb}
\usepackage{mathtools}

\newcommand {\R}{\mathbb{R}}

\newcommand{\set}[1]{\left\{ #1 \right\}}
\newcommand{\citet}[1]{\cite{#1}}

\usepackage{soul}

\usepackage{microtype}
\usepackage{xspace}
\newcommand{\bass}{BASS\textsubscript{ours}\xspace}
\newcommand{\basssrc}{BASS\textsubscript{ours/src}\xspace}
\newcommand{\bassppr}{BASS\textsubscript{ours/ppr}\xspace}
\newcommand{\rtsts}{RTS2S\xspace}
\newcommand{\xrtsts}{exRTS2S\xspace}
\newcommand{\orig}{USG\textsubscript{src}\xspace}
\newcommand{\repro}{USG\textsubscript{ppr}\xspace}

\newcommand{\circled}[1]{{\normalfont \raisebox{.5pt}{\textcircled{\raisebox{-.9pt} {#1}}}}}
\usepackage[frozencache=true,cachedir=minted-cache]{minted}

\usepackage{xcolor}
\definecolor{light-gray}{gray}{0.95}
\newcommand{\code}[1]{\colorbox{light-gray}{\texttt{#1}}}

\fancypagestyle{specialfooter}{%
  \fancyhf{}
  
  \fancyfoot[C]{\vspace{1cm} \hspace*{-.2\textwidth}\parbox{1.4\textwidth}{\small This preprint has not undergone peer review or any post-submission improvements or corrections. The Version of Record of this contribution is published in Advances in Information Retrieval, 46th European Conference on Information Retrieval, ECIR 2024, and is available online at \doi{10.1007/978-3-031-56066-8\_11}.}}
}

\begin{document}
\title{A Second Look on BASS -- Boosting Abstractive Summarization with Unified Semantic Graphs}

\titlerunning{A Second Look on BASS -- A Replication Study}
\author{
\anonymize{
    Osman Alperen Koraş\inst{1,2}\orcidID{0009-0006-6490-3139}
     \and Jörg Schlötterer\inst{3,4}\orcidID{0000-0002-3678-0390} \and
    Christin Seifert\inst{3}\orcidID{0000-0002-6776-3868}
}
{Anonymous Submission}}
\authorrunning{\anonymize{O. A. Koraş et al.}{}}
\institute{\anonymize{Institute for AI in Medicine (IKIM), University Medicine Essen \and University of Duisburg-Essen, Germany
\and
University of Marburg, Germany \and
University of Mannheim, Germany \\
\email{osman.koras@uni-due.de}  \\
\email{\{joerg.schloetterer,christin.seifert\}@uni-marburg.de}
}{}}

\maketitle

\begin{abstract}
\input{src/00_abstract}
\end{abstract}
\thispagestyle{specialfooter}

% This preprint has not undergone peer review or any post-submission improvements or corrections. The Version of Record of this contribution is published in Advances in Information Retrieval, 46th European Conference on Information Retrieval, ECIR 2024, and is available online at https://doi.org/[insert DOI]

\section{Introduction}

\label{sec:introduction}
\input{src/01_introduction}

\section{Replication Methodology}
\label{sec:replication-methodology}
\input{src/02_replication_methodology}

\section{Replicating the BASS framework}
\label{sec:replication-details}
\input{src/03_replication_details}

\section{Evaluation}
\label{sec:experimental-results}
\input{src/05_experiments}

\section{Replication Challenges and Recommendations}
\label{sec:replication-challenges}
\input{src/04_replication_challenges_and_recommendations}

\section{Conclusion}
\label{sec:discussion}
\input{src/06_discussion_and_summary}

\anonymize{
\subsubsection{Acknowledgements}
\input{src/07_acknowledgement}}{}

\bibliographystyle{splncs04}
\bibliography{custom}

\input{src/appendix}

\end{document}

%% file: src/00_abstract.tex
We present a detailed replication study of the BASS framework, an abstractive summarization system based on the notion of Unified Semantic Graphs. Our investigation includes challenges in replicating key components and an ablation study to systematically isolate error sources rooted in replicating novel components. Our findings reveal discrepancies in performance compared to the original work. We highlight the significance of paying careful attention even to reasonably omitted details for replicating advanced frameworks like BASS, and emphasize key practices for writing replicable papers.

\keywords{Replication  \and Abstractive Summarization \and Graph-Enhanced Transformer}

%% file: src/01_introduction.tex
The goal of automatic text summarization is to generate a fluent, concise, informative, and faithful summary of source documents~\cite{ELKASSAS2021113679}. 
{Extractive summarization }systems select salient phrases from the source document and concatenate them to form the summary. 
In contrast, {abstractive summarization }systems freely generate text conditioned on an intermediate representation of the source document~\cite{ELKASSAS2021113679}.
Consequently, the capabilities of abstractive summarization systems depend on the richness of this intermediate representation. 

Many state-of-the-art abstractive summarization systems are based on {Pre-trained Language Models} (PLM), such as {BERT}~\cite{devlin-etal-2019-bert}, {PEGASUS}~\cite{10.5555/3524938.3525989}, or {T5}~\cite{10.5555/3455716.3455856}. And the success of transformers~\cite{10.5555/3295222.3295349} across many domains shows that they are capable of generating rich representations for a wide range of signals, including vision~\cite{Dosovitskiy2020AnII}, audio~\cite{8462506} and graphs~ \cite{Ying2021DoTR}.
The {Graphormer}~\cite{Ying2021DoTR} is one of many Attentive Graph Neural Networks~\cite{Velickovic2017GraphAN,Brody2021HowAA,Ying2021DoTR}, which have been successfully adapted for transformers to leverage graphs in abstractive summarization systems~\cite{10214663,9776298,zhu-etal-2021-enhancing,hu-etal-2021-word,dou-etal-2021-gsum,xu-etal-2020-discourse,huang-etal-2020-knowledge,Jin_Wang_Wan_2020,fan-etal-2019-using}, with the aim to complement or guide the rich representation of transformers with explicitly structured data to improve accuracy and faithfulness of the generated summaries.

One of the graph-enhanced transformer models  is BASS \cite{wu-etal-2021-bass}, which is of specific interest because i) it introduces a compressed dependency graph structure based on the idea of semantic units and ii) the authors report competitive performance in abstractive summarization while being only half the size (201M parameters for {BASS} vs. 406M parameters for {BART} \cite{lewis-etal-2020-bart} and {PEGASUS}).

Because the original paper~\cite{wu-etal-2021-bass} is not accompanied by source code, we conduct a replication study of the BASS framework. Our results contribute to the broader discourse surrounding reproducibility concerns of Machine Learning~\cite{Gibney2022CouldML,Gundersen2022SourcesOI} and in particular NLP research, sometimes even referred to as the ``reproducibility crisis''~\cite{belz-etal-2021-systematic,belz-etal-2023-non}. Belz et. al report that fewer than 15\% scores of their study were reproducible, and that ``worryingly small differences in code have been found to result in big differences in performance''~\citet{belz-etal-2021-systematic}. Even for performance scores reproduced under the same conditions, they discovered that almost 60\% of reproduced scores were worse than the original score.  Consequently, results from different works have to be compared with caution, even if similar components are employed, drawing attention to the importance of generating or replicating own baselines for meaningful comparisons and drawing conclusions.
Concretely, the contributions of this paper are the following:
\begin{enumerate}
    \item We conduct a replication study of the BASS framework and publish our implementation\footnote{\anonymize{https://github.com/osmalpkoras/bass-replication}{To ensure anonymity, we will provide the repository link upon acceptance. Reviewers are directed to the attached ZIP file.}}, including source code for the graph construction component provided by the authors of the original paper.\anonymize{}{\footnote{We thank the authors for their correspondence, their source code and their consent to share it.}}
    \item We conduct an ablation study to examine BASS’ architectural adaptations to incorporate the graph information into transformers and find we can not replicate the performance improvement on the summarization task.
    \item We detail our replication for each framework component, and summarize the specific and general challenges we faced during replication.
\end{enumerate}

%% file: src/02_replication_methodology.tex
\begin{figure}[t]
\includegraphics[width=\textwidth]{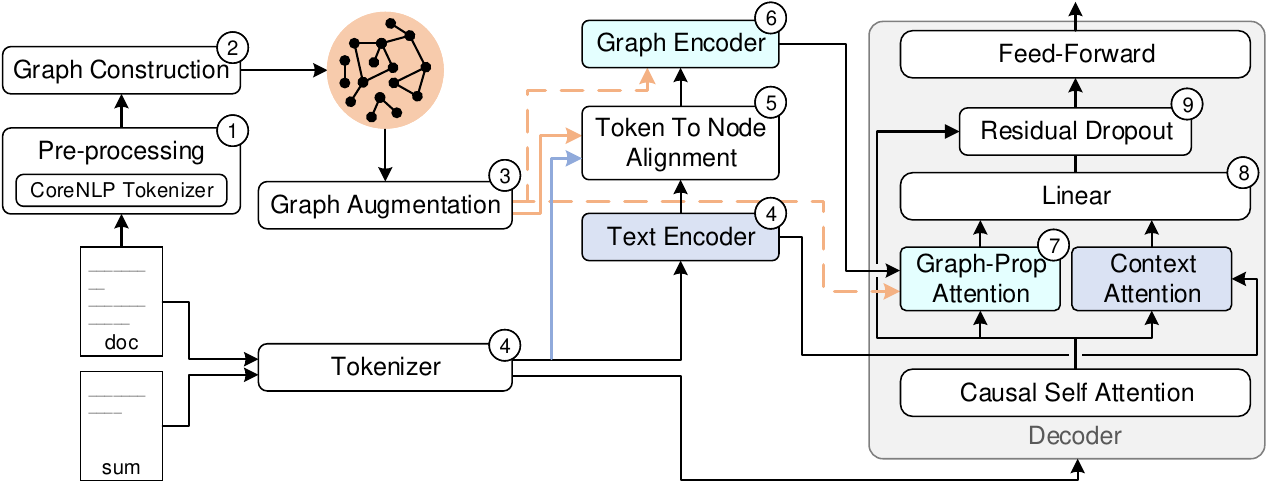}
\caption{Illustration of the BASS framework. The pre-processing and graph construction is done on the input document (left). The resulting graph information is used for token-to-node alignment~\circled{5}, the graph encoder~\circled{6} and the respective cross-attention module~\circled{7} in the decoder.} \label{fig:bass_framework}
\end{figure}

Our initial goal was to implement the BASS framework (cf. Fig.~\ref{fig:bass_framework}) in Python one component at a time, solely from information available to the community, i.e., the paper. We started by implementing the pre-processing~\circled{1} and graph construction~\circled{2}, but quickly identified missing information and uncertainties~(see~Sec.~\ref{sec:replication-challenges}).

When key information was missing on a component, we inquired the authors via email for missing details and source code. When met with uncertainties, we contacted the authors only when we were not certain to have faithfully replicated the component. We exchanged multiple emails with over 20 questions out of which roughly three quarters have been answered. On average, they responded to questions within 10 days.  In the end, the authors provided a Java implementation for the graph construction~\circled{2} and snippets for pre-processing~\circled{1}. But we were not provided with information about the batch size for example, despite multiple inquiries.

With the additional information, we found some inconsistencies between author information, paper details and source code (see Sec.~\ref{subsec:graph-construction} and Appendix~\ref{sec:code-comparison}). We let details provided in the paper take precedence over implementation details of the Java source code, and let the source code take precedence over our correspondence with the authors. For details that still remained unclear, we made a best guess.

A summary of our replication is shown in Tab.~\ref{tab:replication-overview}, where we indicate per component which information sources we used, and whether uncertainties remained. We implemented  the pre-processing~\circled{1} in Java to use the authors' source code for graph construction~\circled{2}a, but additionally replicated the graph construction~\circled{2}b in Python. All other components~\circled{3}~--~\circled{9} are replicated in Python only. To switch between programming languages, we save the pre-processing and graph construction output as needed for subsequent computation in Python.

\begin{table}[t]
\caption{Overview of the completeness of information on components (cf. Fig.~\ref{fig:bass_framework}) based on community-available information, i.e., the paper, and which information \textbf{Sources} we actually used. \textbf{Paper} shows whether key information was missing, making a replication impossible ($\times$), whether minor details were omitted (\circled{}) or whether all required information on a component is complete (\checkmark).
\textbf{Complete} shows whether we are certain to have faithfully replicated a component (\checkmark) or if uncertainty remained (\circled{}). n.a.: not applicable to graph construction~\circled{2}a,~as~we~use~the~provided~source~code~as~it~is.}
\label{tab:replication-overview}
\begin{tabular}{+l^c^l^c}
\toprule\tabhead 
Component &  Paper  & Sources &  Complete \\\otoprule
\circled{1} Pre-processing   & $\times$ & Paper, Source Code, Authors & \checkmark  \\
\circled{2}a Graph Construction (provided) & n.a. & Source Code & n.a.  \\
\circled{2}b Graph Construction (replicated) & $\times$ & Paper, Source Code, Authors & \circled{} \\
\circled{3} Graph Augmentation & \checkmark & Paper & \checkmark \\
\circled{4} Tokenizer \& Text Encoder & \circled{} & Paper, Authors & \checkmark \\
\circled{5} Token To Node Alignment & \circled{}  & Paper & \checkmark \\
\circled{6} -- \circled{9} Model Architecture & \circled{} & Paper, Authors & \circled{} \\

\bottomrule
\end{tabular}
\end{table}

%% file: src/03_replication_details.tex
BASS is an abstractive summarization framework (cf. Fig.~\ref{fig:bass_framework}), which uses i) dependency parse trees to generate \textit{Unified Semantic Graphs} (USG) for documents to compress and relate information across the input document, and ii) a model architecture, which incorporates the graph information. As this work focuses on the replication of the framework, we refer to the original work~\cite{wu-etal-2021-bass} for it's details and only elaborate on necessary complementary~information~in~the~following.

\paragraph{Pre-processing~\circled{1}.}
An input document is passed to a linguistic parser for POS tagging, co-reference resolution and dependency parsing. We used the latest CoreNLP library~\cite{manning-EtAl:2014:P14-5} (v4.5.2) and had the authors confirm a configuration we found in their source code, as no details were given in the paper, that is:
\begin{minted}[fontsize=\small,breaklines, breakautoindent]{json}
"annotators": "tokenize, ssplit, pos, lemma, ner, parse, coref",
"coref.algorithm": "neural", "depparse.extradependencies": "MAXIMAL"
\end{minted}

\noindent We were initially unable to pre-process many documents (over 20\%) due to endless runtimes or out-of-memory errors of the CoreNLP parser. Upon inquiry, the authors' confirmed to have used the pre-processing strategy indicated in their source code, so we chunk source documents into blocks of sentences with approx. 500 words and pre-process them separately. The resulting graphs per chunk are concatenated into a single document-level graph consisting of multiple disjoint sub-graphs. In particular, nodes across different sub-graphs (e.g., those referring to the same entity) are neither merged nor connected. We additionally set a maximum runtime (of up to 10 hours) and RAM consumption (of up to 90GB) for pre-processing a single chunk. The output is one dependency parse tree per sentence, co-reference chains across each chunk and a POS tag for each word.

\paragraph{Graph Construction~\circled{2}.}
\label{subsec:graph-construction}

To construct USGs, the dependency trees of all sentences are viewed as directed graphs with one node representing a single word, paired with its POS tag and an edge representing the respective dependency relation. Nodes are merged in multiple steps as follows: 
i) the nodes of entity mentions from co-reference resolution are merged into a single entity node,
ii) nodes are removed or merged based on linguistic rules reasoning over the dependency and POS annotations, 
iii) entity nodes of mentions in the same co-reference chain are merged,
iv) non-entity nodes representing the same phrase (exact lexical match) are merged. 

We noticed strictly complying with the paper \cite{wu-etal-2021-bass} yields a different algorithm than the one provided with the Java source code  (Appendix~\ref{sec:code-comparison}). Therefore, we consider two USG variants in this work: i) \orig generated by original Java implementation~\circled{2}a and ii) \repro generated by our paper-compliant replication of the graph construction~\circled{2}b in Python. Since the linguistic rules in step iii) are omitted from the paper, we adopt these from the Java code.

\paragraph{Graph Augmentation~\circled{3}.}
Following the paper, the USG is augmented by adding self-loops and reverse edges. All nodes are connected with their two-hop neighbours and a supernode connecting to all nodes is added. The output of this step is two matrices, the graph construction matrix $C$ (solid orange line) and the adjacency matrix $A$ (dashed orange line). See Appendix~\ref{sec:aligning-graph-and-text-embeddings} for more details.

\paragraph{Text Encoder and Tokenizer~\circled{4}.}
Following our correspondence with the authors, we use the pre-trained RoBERTa-base model \cite{zhuang-etal-2021-robustly} for the text encoder with the tokenizer \code{RobertaTokenizerFast} from Hugging Face's transformer library (\citet{wolf-etal-2020-transformers}, v4.26.1). The context length is extended to $N = 1024$ by randomly initializing the extended part.

\paragraph{Aligning Text to Graph Embeddings~\circled{5}.}
We align text embeddings with graph embeddings by multiplying the text encoder output with the graph construction matrix from step~\circled{3}. 
We refer to Appendix~\ref{sec:aligning-graph-and-text-embeddings} for more details.

\paragraph{Model Architecture~\circled{6} --~\circled{9}.}

The model architecture builds upon a standard transformer encoder-decoder architecture for abstractive summarization, complemented with three additional components: 
i) a graph encoder~\circled{6}, which is a standard two-layered transformer encoder using the adjacency matrix of the graph as attention mask, 
ii) a corresponding multi-head attention module~\circled{7} in a six-layered decoder attending over the graph encoder output and 
iii) a fully connected linear layer~\circled{8} to fuse graph and text information, followed by a Residual Dropout~\cite{10.5555/3295222.3295349} layer~\circled{9}.

The Causal Self Attention, i.e., masked self attention attending to left context only, and Feed-Forward~\cite{10.5555/3295222.3295349} layers are also followed by a Residual Dropout layer, not indicated in the figure.  In contrast to regular cross-attention modules, the attention weights in~\circled{7} are propagated using PageRank and the augmented adjacency matrix to leverage the graph structure. We refer to Appendix~\ref{sec:graph-propagation} for more details on our implementation.  In the end, our model ended up having approx. 205M trainable parameters, which is around 2\% larger than reported in the BASS paper~\citet{wu-etal-2021-bass} (201M parameters), implying architectural differences we could not entirely resolve.

%% file: src/05_experiments.tex
We conduct a replication study by training one model each for the author-provided graphs \orig and the paper-compliant graphs \repro (cf. Sec.~\ref{subsec:graph-construction}) to better isolate potential errors rooted in replicating the model architecture from potential errors rooted in replicating the graph construction. Since replication studies are known to be challenging due to an overwhelming number of error sources, we follow up with an ablation study where we generate our own baselines (not using graphs) to measure the impact of architectural adaptations.

\paragraph{Datasets.} We follow~\citet{wu-etal-2021-bass} and select the {BigPatent}~\cite{sharma-etal-2019-bigpatent} dataset for {Single Document Summarization}. However, we were only able to pre-process $99.79\%$ documents, resulting in 1,204,631 documents for training  and 66,962 documents each for validation and test. Using both graph construction methods (cf.~Sec.~\ref{subsec:graph-construction}) we generate two sets of graphs \orig and \repro and obtain two respective training datasets. Running the pre-processing and graph construction~\circled{2}a on the Big Patent dataset took about four days on our cluster with 900 CPU cores (Intel(R) Xeon(R) Silver 4216 CPU) with an aggregated runtime of 3,360~days~$\pm$~1~day.

\paragraph{Models.}
We use the following baselines for our studies (cf.~\cite[Tab.3]{wu-etal-2021-bass} for original results):
\textbf{TransS2S}, which is a standard text-only transformer encoder-decoder model~\cite{10.5555/3295222.3295349}, 
\textbf{RoBERTaS2S}, which differs from TransS2S by using RoBERTa-base for the encoder and decoder, and \textbf{BASS}, which is the  original model we replicate. 

For our replication study, we train the following models:
\textbf{\basssrc}, which is our replicated BASS model trained with \orig, and \textbf{\bassppr}, which is a full replication  of the original paper differing from \basssrc only by it's use of \repro graphs.

For the ablation study, we additionally train two transformer based encoder-decoder architectures on the BigPatent dataset without any graphs:
\textbf{\rtsts}, which is a text-only model consisting of the text encoder~\circled{4} and a standard 6-layered decoder without any graph components (i.e., omitting~\circled{3},\circled{5}~--~\circled{8}). And
\textbf{\xrtsts}, which extends \rtsts by the graph encoder and the decoder components~\circled{6}~--~\circled{8} without informative graph structure (\circled{3},\circled{5}), and by replacing Graph-prop Attention~\circled{7} with normal Context Attention.

\rtsts is most similar to TransS2S and RoBERTaS2S, using the encoder of RoBERTaS2S and the decoder of TransS2S. We therefore expect this model to perform somewhere in-between. \xrtsts is most similar to \bass{} and differs only in the lack of graph structure: every token is considered a graph node connected to every other node. Hence layer~\circled{5} is skipped and no graph structure is injected into the attention mechanism in the graph-encoder~\circled{6} and no attention is graph-propagated in the cross-attention module~\circled{7}.

We want to measure the impact of all architectural adaptations proposed for BASS with \rtsts as our baseline. With \xrtsts as a baseline, we further isolate the impact of the \orig structure from the impact of increasing model size. We choose \orig assuming the graph construction method~\circled{2}a reproduces the graphs from the original work.

\paragraph{Training Details.}
We use the same training and hyper-parameter setup as the original work~\cite[\S5.2]{wu-etal-2021-bass} for all models, which uses the learning rate schedule of Liu and Lapata~\citet{liu-lapata-2019-text}. Each model is trained once for 300,000 steps. 
Since we were not able to find out the batch size used in the original work, we use a batch size of 48 per step, which is the largest possible value on our hardware (6 RTX A6000 GPUs with a total RAM of 288GB -- the original authors reported the use of 8xV100, of which the latest version totals 256GB). Training the models for 10,000 steps took about 4 hours $\pm$ 30 minutes at average on our machines~(cf.~Tab.~\ref{tab:results}).

\paragraph{Evaluation.}
We evaluate the models on the test set and apply beam search \cite{DBLP:journals/corr/WuSCLNMKCGMKSJL16} with trigram blocking \cite{paulus2017deep} for decoding using a beam size of $5$ and a length penalty of $0.9$. We enforce a maximum decoding length of 1024 and report ROUGE  scores~\cite{rougescore} R-1, R-2, sentence-level R-L, summary-level R-L$_{sum}$ and $F_1$ BERTScore~\cite{bert-score}\footnote{\code{lang="en"} \code{model\_type="roberta-large"} \code{rescale\_with\_baseline=True}} BS. We exclude summaries, for which no \code{eos} token has been generated during decoding from evaluation and use paired bootstrap resampling~\cite{dror-etal-2018-hitchhikers} with $p=0.05$ for significance testing.

\subsection{Experimental Results}
\paragraph{Replication.}
% BASS MODELS
Comparing our replication with the original (\textbf{BASS}), we score more than 4 points lower than reported originally,\textsuperscript{\ref{ftnt:results}} performing even worse than the RoBERTaS2S baseline. % core!
% EXTENDED TRAINING
Since the training batch size of the original work remains unknown, we investigated if our models might be undertrained by \textbf{extending the training} of \basssrc for another 300,000 steps\footnote{We did not train further, because we already doubled the computational budget used for the original paper.} (cf. Tab.~\ref{tab:results}) and are indeed able to raise the scores but only by about $0.7 \pm 0.2$ points.

% GRAPH STRUCTURE
Seeing how the graph construction implementation~\circled{2}a provided to us differs algorithmically from our paper-compliant implementation~\circled{2}b, we also compare \orig and \repro \textbf{graphs} (cf. Tab.~\ref{tab:comparing-graph-structures}). The replicated graphs are slightly larger in size. They also cover at least 30\% more tokens and while we expected \bassppr to perform better for this reason, we actually observe mixed results: a small increase in the BERTscore, and a decrease in ROUGE scores.

\input{src/table-results}
\input{src/table-graph-comparison}
\paragraph{Ablation.}
We observe slight but mostly significant differences in model performances (cf. Tab.~\ref{tab:results}, {\sc Ablation Study}), with no clear winner. However, our ablation models perform in-between TransS2S and RoBERTaS2S, as expected. The introduction of graph components (\xrtsts vs. \rtsts) mostly improves the BERTScore with mixed results in terms of ROUGE. Further comparing \xrtsts with our replicated models shows the impact of using USGs: using \repro slightly hurts performance overall, while using \orig slightly improves ROUGE scores while hurting BERTScore.

\subsection{Discussion}
\label{subsec:discussion}
\paragraph{Replication.}
% BASS MODELS
Our replicated \textbf{BASS} models substantially fall short in performance, even below baselines of the original work (cf. RoBERTaS2S in Tab.\ref{tab:results}). Since this is true even for our model \basssrc which is trained on \orig graphs, we mainly attribute this to the model architecture, assuming that the provided graph construction \circled{2}a is the same one used in the original work.
% EXTENDED TRAINING
By further \textbf{extending the training}, we find that our models might be undertrained, indicating that the original work might have used a larger effective batch size to achieve their results.

% GRAPH STRUCTURE
Since the discrepancy between the replicated and the original performance of BASS can be attributed to the model architecture, the impact of using \repro over \orig \textbf{graphs} on downstream performance being minor does not surprise, despite substantial qualitative differences in graph structures. However, the authors' source code not complying with the paper, possibly having undergone some changes (see Sec.~\ref{sec:replication-details}~\&~\ref{sec:replication-challenges}), sheds doubts on whether the provided graph construction \circled{2}a reproduces the graphs of the original work and whether the mismatch in graph structures indicates a failed replication of the original work.

\paragraph{Ablation.}
Contrary to \cite{wu-etal-2021-bass} we did not find substantial performance gains, neither in model adaptations (increased model size, cf. Tab.~\ref{tab:results}, \rtsts vs. \xrtsts), nor in USGs (additional structured information, cf. Tab.~\ref{tab:results}, \xrtsts vs. \basssrc and \bassppr). This is shown by the minor impact the model adaptations and USGs have on our baselines. Our replicated \repro graphs even consistently hurt the performance overall. Nonetheless the comparison with previous baselines (TransS2S and RoBERTaS2S) shows that \rtsts performs reasonably well. We therefore ascribe the lack of substantial gains in our replicated models solely to BASS' model adaptations and graph information not being as effective as expected.

%% file: src/table-results.tex
\begin{table}[t]
    \centering
    \caption{Evaluation of our models on the BigPatent dataset. The baselines are all taken from prior work~\cite{wu-etal-2021-bass} and best scores are in bold. All scores are pairwise significantly different from each other, except those indicated by $\dagger$.}
    \begin{tabular}{+l@{\hspace{0.3cm}}^c@{\hspace{0.2cm}}^c@{\hspace{0.2cm}}^c@{\hspace{0.2cm}}^c@{\hspace{0.2cm}}^c@{\hspace{0.2cm}}^c@{\hspace{0.2cm}}^c}
    \toprule \tabhead
     & R-1 & R-2 & R-L & R-L$_{sum}$ & BS & params & time \\\otoprule
     \multicolumn{8}{c}{\sc  Baselines and Paper}\\
     TransS2S \cite{wu-etal-2021-bass}  & 34.93 & 9.86 & \multicolumn{2}{c}{ 29.92\tablefootnote{While the BASS paper reports sentence-level R-L scores, they systematically match better with our summary-level R-L$_{sum}$ scores, which may indicate that prior results are actually R-L$_{sum}$ scores. Hence we place the scores reported by BASS between columns and compare them with our R-L$_{sum}$ scores.\label{ftnt:results}}}\hspace{0.2cm} & 9.42 & n.a. & n.a.\\
    RoBERTaS2S \cite{wu-etal-2021-bass}  & 43.62 & 18.62 & \multicolumn{2}{c}{37.86\tablefootnotemark{ftnt:results}}\hspace{0.2cm}  & 18.18 & n.a.  & n.a.\\
    BASS \cite{wu-etal-2021-bass}  & \textbf{45.04} & \textbf{20.32 }& \multicolumn{2}{c}{ \textbf{39.21}\tablefootnotemark{ftnt:results}}\hspace{0.2cm}  & \textbf{20.13} & 201M & n.a. \\
    \multicolumn{8}{c}{\sc  Replication Study}\\
    \basssrc & {40.52} & {14.97} & {26.64} & {34.67} & 16.15 & 204.6M & 5d 15h \\
    \basssrc\textsubscript{600k} & \textbf{41.23} & \textbf{15.62 }& \textbf{27.18} & \textbf{35.37} & \textbf{17.03} & 204.6M & 11d 11h \\
    \bassppr & 39.30 & 14.65 & 26.31 & 33.74 & 16.51 & 204.6M & 5d 16h \\
    \multicolumn{8}{c}{\sc  Ablation Study}\\
    \rtsts{} & \textbf{39.75}  & \textbf{14.84$^\dagger$} & 26.48 & \textbf{34.04} & 16.34 & 172.4M & 4d 9h \\
    \xrtsts{} & {39.62}& \textbf{14.86$^\dagger$} & \textbf{26.56}& {33.91} & \textbf{16.63} & 204.6M & 5d 7h \\
    \bottomrule
    \end{tabular}
    \label{tab:results}
\end{table}

%% file: src/table-graph-comparison.tex
\begin{table*}[t]
    \centering
    \caption{Comparison of \orig and \repro structures for the BigPatent dataset $D = \set{d_i}_{i\in I}$, with $d_i$ denoting a tokenized input document. We consider the subsets $D_T = \set{d_i \mid \left|T - t(d_i)\right| \leq 20}$ for token count $t(d_i)$ and $T \in \set{400,600,800,1000}$. Let $\overline{t}_{D_T}$ denote the average token count of documents $d \in D_T$ and $\left|D_T\right|$ the cardinality of $D_T$. We report the average node count $\overline{n}$, average edge count $\overline{e}$ as well as the average count of tokens $\overline{t}_c$ covered by the graphs generated for $d \in D_T$. The bottom row shows the increase in respective quantities for \repro w.r.t \orig}
    \begin{tabular}{+c@{\hspace{0.3cm}}c^c^c@{\hspace{0.3cm}}c^c^c@{\hspace{0.3cm}}c^c^c@{\hspace{0.3cm}}c^c^c}
    \toprule \tabhead
     $\overline{t}_{D_T}$ &   & 400 &  &  & 600 &  &  & 800 &  &  & 1000 & \\
     $\left|D_T\right|$ &   & 152 &  &  & 966 &  &  & 3045 &  &  & 6087 & \\
     & $\overline{n}$ & $\overline{e}$ & $\overline{t}_c$  & $\overline{n}$ & $\overline{e}$& $\overline{t}_c$  & $\overline{n}$ & $\overline{e}$& $\overline{t}_c$  & $\overline{n}$ & $\overline{e}$& $\overline{t}_c$ \\\otoprule
    \orig & 117 & 142 & 232 & 168 & 211 & 340 & 227 & 281 & 463 & 278 & 349 & 569 \\
    \repro &	129 & 156 & 301 & 185 & 233 & 453 & 240 & 308 & 606 & 293 & 385 & 759 \\\midrule
     Increase &	10\% & 10\% & 30\% & 10\% & 10\% &33\% & 6\% & 10\% & 31\% & 5\% & 10\% & 33\%
    \\\bottomrule
    \end{tabular}
\label{tab:comparing-graph-structures}
\end{table*}

%% file: src/04_replication_challenges_and_recommendations.tex
In this section, we reflect on the replication process and highlight the main challenges we encountered, hoping to sensitize readers to the underlying issues that compromise the replicability of research papers.  We conclude by recommending key practices for writing replicable papers, that would have significantly helped us with the replication.

\paragraph{Self-Explanatory Details.} Some details are omitted from papers usually for good reasons:  being straight-forward, well-known or trivial. 
However, our experience showed that the leeway in implementation choices for omitted details (e.g. in every step of the graph construction, or for aligning the tokenizations of CoreNLP's tokenizer~\circled{1} and RoBERTa's tokenizer~\circled{4},  but also for the pre-processing strategy) entails ambiguities and consequently an avalanche of potential error sources and mitigation strategies to resolve them. Hence, these omitted details can make the difference between an accurate replication and an endless errand to fix errors. 
Although we got many details confirmed by the authors, studied the source code and strictly followed the paper, some uncertainties remained (cf. Tab.~\ref{tab:replication-overview}). In the end, we were neither able to pre-process the entire dataset considered in this work (see Sec.~\ref{sec:experimental-results}) nor achieve the same model size as the original authors (see~Sec.~\ref{sec:replication-details}).

\paragraph{Missing Third Party Information.} One problem was missing version information and configuration of third-party components, i.e., of the CoreNLP pipeline, the RoBERTa model and the tokenizers. We were able to resolve most of these issues through correspondence with the authors for this paper.

\paragraph{Missing Key Information.} Overall, we encountered many details that required additional information or clarification beyond the paper. However, not all missing details fundamentally obstructed our replication: our first attempt to replicate the paper (see Sec.~\ref{sec:replication-methodology}) failed primarily due to the omission of the linguistic rules reasoning over the dependency trees for creating USGs. Nevertheless, following our correspondence with the authors and access to source code, we identified and resolved many misunderstandings. 

\paragraph{Algorithmic Complexity and Error-Proneness.} A thorough analysis of the original source code was necessary to fix (or not to fix) our replicated graph construction algorithm due to the many errors we encountered during runtime, often rooted in erroneous annotation results, such as wrong POS tag annotation, co-reference resolution, or even dependency graphs being rooted in punctuation tokens or sentences mistakenly being split at decimal points. To our surprise, the provided graph construction slightly differs from the description in the paper.  This is likely because the source code has been used in other projects, as noted by the authors, and consequently might have undergone some changes before or after the paper's publication, emphasizing the importance of version control systems. The complexity of algorithms, whereas,  can be lessened using tools during development to analyze and reduce cognitive complexity in software.

\paragraph{Recommendations.} We found the mathematical and algorithmic descriptions (notation, equations, pseudo-code) most helpful along the way, allowing us to consolidate many misconceptions. Therefore we emphasize the importance of 
i) providing a clear and complete technical context,
ii) a clear and (given the context) complete notation,
iii) technical and mathematical precision particularly for describing how different components (novel or not) interleave,
and iv) commented pseudo-code. We feel the latter can often replace a detailed description of an algorithm, while being shorter and less ambiguous. 

We also strongly encourage to use technical terms coined by prior work  wherever applicable, such as ``Residual Dropout''~\cite{10.5555/3295222.3295349} for layer~\circled{9}, instead of short descriptions of well-known components. The latter can be inaccurate and leave readers questioning potential differences and misunderstandings in case of failed replications, while undermining the development of a well-defined and well-known terminology of a research domain.

%% file: src/06_discussion_and_summary.tex
% CONCLUSION REPLICATION
We started implementing the BASS framework based on the paper, but found that most components were not sufficiently described (cf. Tab.~\ref{tab:replication-overview}). Some uncertainties persisted even after our correspondence with the authors, and the examination of the provided source code (cf. Sec.~\ref{sec:replication-methodology}), partly because some inquiries (e.g. for the training batch size) had been left pending. On one hand, the provided graph construction \circled{2}a did not align with the paper (cf. Sec.~\ref{subsec:graph-construction}). On the other hand, our model's parameter size was larger by approx 3.6M parameters than reported in the original work (cf. Sec.~\ref{subsec:graph-construction}). Therefore, it is unsurprising we could neither replicate prior results of BASS on the BigPatent dataset, nor clear performance improvements on the summarization task as a result of the novel adaptations proposed for BASS (cf. Sec.~\ref{subsec:discussion}). Assuming the graph construction method \circled{2}a provided by the original authors' reproduces the same USGs as in the original work, our results indicate the poor performance can be ascribed to the model architecture, which might, in addition, be undertrained. However, some doubts remain on whether the provided graph construction method even reproduces the original USGs.

% CONCLUSION UNIFIED SEMANTIC GRAPHS
Moreover, we found the pre-processing~\circled{1} and in extension the graph construction~\circled{2} to be very error-prone and time consuming. Parsing one document of the BigPatent dataset with approx. 1,000 tokens took us about 3.5 minutes, not accounting for the 2,811 documents (approx. 0.2\%) we had to exclude for not being parseable in less than 10 hours. Additionally, erroneous dependency annotations make it difficult to construct USGs, leading to fractured graphs, isolated nodes or deletion of salient information. Based on our experiences, we suggest investigating the use of simpler semantic dependency parsing methods~\cite{dozat-manning-2018-simpler} which reportedly are more accurate, or to move away from systems that construct graphs from semantic annotations based on manually hand-crafted rules.

Overall, the replication was complicated by missing third-party information, the ambiguity of self-explanatory details, and omission of some key information (the latter requiring us to contact the authors for a faithful replication), despite the fact that the original paper is very detailed and comprehensive, representative of the high quality of the venue it was published on (ACL'21). However, our experience shows that the way information is detailed is just as vital as being comprehensive. Furthermore, as this lesson is learned only after attempting a replication, it may lead reviewers, who lack similar experiences, to overrated reproducibility assessments. We have therefore emphasized key issues and practices for replicable papers and recommend supplementing reproducibility as well as reviewer checklists with a corresponding section to address these problems.

%% file: src/07_acknowledgement.tex
We thank the authors~\cite{wu-etal-2021-bass} for their correspondence, their source code and for giving us their consent to share it. Funding for this work was provided by the German State Ministry of Culture and Science NRW, for research under the Cancer Research Center Cologne Essen (CCCE) foundation. The funding was not provided specifically for this project. 

%% file: src/appendix.tex
\appendix
\section{Discrepancies between \orig{} and \repro{}}
\label{sec:code-comparison}
In the following, we point out algorithmic differences between the two graph constructions methods \circled{2}a and \circled{2}b (cf. Tab.~\ref{tab:replication-overview}). These differences arise from strictly complying with the paper for the replicated method \circled{2}b. For the sake of simplicity, we align our comparison with the pseudo-code in the appendix of the original work~\cite{wu-etal-2021-bass}.

\paragraph{\code{REMOVE\_PUNCTUATION}} While \circled{2}b removes all tokens whose dependency relation equals \code{punct} or whose POS tag is an element of $P = $\{{\code{.}, \code{,}, \code{:}, \code{!}, \code{?}, \code{(}, \code{)}}\}, \circled{2}a only removes tokens whose POS tag is an element of $P$. In \circled{2}a, punctuation is removed recursively as part of the \code{MERGE\_NODES} routine.

\paragraph{\code{MERGE\_COREF\_PHRASE}}
\circled{2}b merges all tokens of a co-reference mention into a single node, while \circled{2}a i) uses merging rules not mentioned in the paper and also ii) immediately merges the resulting nodes in the same co-reference chain into a single node. \circled{2}b, on the other hand, merges nodes in the same co-reference chain only in the \code{MERGE\_PHRASES} step.

\paragraph{\code{MERGE\_NODES}} While \circled{2}a and \circled{2}b use exactly the same rules to merge nodes, \circled{2}a traverses the dependency trees pre-order depth-first. We intuitively chose to traverse post-order depth-first for \circled{2}b without looking at the provided source code, as working through the tree bottom up from leaf to root nodes generally complies better with the intention to merge nodes (which includes rules to delete nodes). For example, if \circled{2}a deletes a child, all descendants are detached from the tree and never visited by the algorithm.

\paragraph{\code{MERGE\_PHRASES}} \circled{2}b merges the nodes of mentions in the same co-reference chain into a single node and later merges all nodes, whose phrases are equal (exact lexical match). \circled{2}a, on the other hand, only does the latter.

\section{Aligning Graph and Text Embeddings}
\label{sec:aligning-graph-and-text-embeddings}

The Unified Semantic Graph imposes a graph structure on the token embeddings returned by the text encoder. The text encoder output $t$ must therefore be mapped to the graph encoder input $g$, before passing it to the graph encoder. For this, we match tokens with nodes based on text characters from left to right, as the CoreNLP tokenizer~\circled{1} is different from RoBERTa's tokenizer~\circled{4}.

Let $G := (V, E)$ be the augmented Unified Semantic Graph~\circled{3} with nodes $V := \set{v_i}_{i=0}^{N_V}$ and edges $E := \set{e_{ij}} \subset V \times V$ and let $S = \set{s_{v_i}}_{i=0}^{N_V}$ with $s_{v_j} = \set{c_i}_{i \in I_{v_j}}$ be the set of characters of the input document $D = \set{c_i}_{i=0}^{N_D}$ being represented by node $v_j$. As a result of merging nodes across a document, $s_{v_j}$ may consist of multiple disconnected character sequences.

We map nodes $v_j$ to tokens $t_{v_j} \subset T  $, where $t_{v_j} $ is the subset of input tokens $T = \set{t_i}_{i=0}^{N_T}$ associated with at least one character $c_i \in s_{v_j}$. This gives us the graph construction matrix $C = (c_{ij}) \in \R^{N_V \times N_T}$ and the adjacency matrix $A = (a_{ij}) \in \R^{N_V \times N_V} $ with 

\vspace{\abovedisplayskip}
\begin{minipage}[c]{0.45\textwidth}

$$c_{ij} = 
\begin{cases}
    1, & \text{if } t_j \in t_{v_i}\\
    0, & \text{otherwise}
\end{cases}$$
\end{minipage}
\begin{minipage}[c]{0.45\textwidth} 
$$a_{ij} = 
\begin{cases}
    1, & \text{if } e_{ij} \in E\\
    0, & \text{otherwise.}
\end{cases}$$
\end{minipage}
\vspace{\belowdisplayskip}

Let $d_{model}$ be the dimension of token embeddings and $t \in \R^{N_T \times d_{model}}$ be the output of the text encoder. The graph encoder input $g$ is then given by $g := C't$ where $C'$ is the degree normalized graph construction matrix $C$. Multiplicating $t$ with $C'$ is equal to calculating the representation of node $v_j$ by averaging over the tokens $t_{v_j}$. The matrix $g$ is then passed to the graph encoder alongside the node padding and the adjacency matrix $A$ as attention mask.

\section{Graph Propagation}
\label{sec:graph-propagation}
The paper suggests propagating attention weights in the cross attention module for the graph encoder using PageRank \cite{Klicpera2018PredictTP} and the adjacency matrix $A$ given by the augmented Unified Semantic Graph.  For this, we compute the graph propagation matrix 
$P = \omega^p \hat{A} + (1-\omega)(\sum_{i=0}^{p-1} \omega^i \hat{A}^i)$,
where $p$ is the number of aggregation steps, $\omega$ is the teleport probability and $\hat{A}$ is the degree normalized adjacency matrix $A$, including self loops and reverse edges, supernode edges and shortcut edges. The graph propagated attention weights are then computed as $\alpha' = \alpha P^T$, where $\alpha = (\alpha_{ij}) \in \R^{N_V \times N_V}$ are the attention weights of a single head in the multi-headed cross-attention module for the graph encoder given by 
$\alpha_{ij} = (y_i W_Q) (v_j W_K)^T / \sqrt{d_k}$,
with query and key projection weights $W_Q, W_K$, the $i$-th token representation as query $y_i$ and the $j$-th node representation as key $v_j$. Here, $d_k$ denotes the query and key dimensions.